\documentclass[conference]{IEEEtran}
\IEEEoverridecommandlockouts                              

\usepackage{blindtext}
\usepackage{tikz}
\usepackage[utf8]{inputenc}
\usepackage[english]{babel}
\usepackage{subcaption}  

\usepackage[ruled,vlined,linesnumbered]{algorithm2e}
\usetikzlibrary{positioning}
\usetikzlibrary{shapes.geometric}
\usetikzlibrary{arrows.meta}
\usepackage[nolist]{acronym}
\usepackage{array} 
\usepackage{amssymb} 
\usepackage{amsmath}  %

\usepackage{booktabs}
\graphicspath{{figures/}} 
\DeclareGraphicsExtensions{.pdf,.jpg,.png} 
\usepackage{amsfonts} 
\usepackage{mathtools} 
\usepackage{tikz} 
\usetikzlibrary{shapes,arrows,fit,calc,positioning,automata,backgrounds}
\usepackage{color} 
\usepackage{multirow} 
\usepackage{hhline} 
\usepackage{booktabs}
\usepackage[bookmarks=false, breaklinks]{hyperref} 
\usepackage{enumerate} 
\usepackage{epstopdf} 
\usepackage{dsfont}
\usepackage{siunitx} 
\usepackage{cite}   
\usepackage{diagbox}    
\usepackage{pdfpages}   
\usepackage{breqn}
\usepackage{svg}
\usepackage{multicol}
\newtheorem{remark}{Remark}

\svgpath{{images/svg/}} 
\colorlet{Green1}{green!90!}
\colorlet{Green2}{green!60!}
\colorlet{Green3}{green!40!}
\colorlet{Green4}{green!20!}
\colorlet{Green5}{green!10!}

\usepackage{url}

\usepackage{breakurl}

\RequirePackage{tikz} 

\tikzstyle{startstop} = [rectangle, rounded corners, minimum width=3cm, minimum height=1cm, text centered, draw=black, fill=blue!20]
\tikzstyle{process} = [rectangle, minimum width=3cm, minimum height=1cm, text centered, draw=black, fill=green!20]
\tikzstyle{decision} = [diamond, minimum width=3cm, minimum height=1cm, text centered, draw=black, fill=red!20]
\tikzstyle{arrow} = [thick,-{Latex[length=3mm, width=2mm]}]

\definecolor{Bookcolor}{HTML}{00F9DE}
\definecolor{darkgreen}{rgb}{0.0, 0.5, 0.0}
\definecolor{gray}{gray}{0.9}
\definecolor{lightgray}{rgb}{0.86, 0.86, 0.86}

\makeatletter
\def\@citex[#1]#2{\leavevmode
\let\@citea\@empty
\@cite{\@for\@citeb:=#2\do
{\@citea\def\@citea{,\penalty\@m\ }%
\edef\@citeb{\expandafter\@firstofone\@citeb\@empty}%
\if@filesw\immediate\write\@auxout{\string\citation{\@citeb}}\fi
\@ifundefined{b@\@citeb}{\hbox{\reset@font\bfseries ?}%
\G@refundefinedtrue
\@latex@warning
{Citation `\@citeb' on page \thepage \space undefined}}%
{\@cite@ofmt{\csname b@\@citeb\endcsname}}}}{#1}}
\makeatother

\begin{acronym}
{
\acroplural{ROV}[ROVs]{Remotely operated vehicles}
    \acro{MPC}{model predictive control}
    \acro{GP}{Gaussian process}
    \acro{AUV}{autonomous underwater vehicle}
    \acro{MHE}{moving horizon estimator}
    \acro{EKF}{extended kalman filter}
    \acro{ROV}{remotely operated vehicle}
    \acro{ROVs}{remotely operated vehicle}

    \acro{TSP}{traveling salesman problem}
    \acro{IMU}{inertial measurement unit}
    \acro{DVL}{doppler velocity log}
    \acro{NED}{North East Down}
    \acro{RTI}{real-time iteration}
    \acro{OCP}{optimal control problem}
    \acro{LP}{linear program}
    \acro{MIP}{mixed integer program}
    \acro{EA-MPC}{entanglement aware model predictive control}
    \acro{TSDF}{ truncated signed distance field}
    \acro{SDF}{ signed distance field}
    \acro{CPP}{ coverage path planner}
    \acro{OEA-PP}{online entanglement aware path planner}
    \acro{OMPL}{open motion planning}
        \acroplural{ROV}[ROVs]{remotely operated vehicles}

\acro{REACT}{real-time entanglement-aware coverage path planning for tethered underwater vehicles}

}

\end{acronym}

\begin{document}
\title{\textbf{REACT}: \textbf{Real-time Entanglement-Aware Coverage Path Planning for Tethered Underwater Vehicles}}

\author{Abdelhakim Amer, Mohit Mehndiratta, Yury Brodskiy, Bilal Wehbe, and Erdal Kayacan%
\thanks{A. Amer is with the Artificial Intelligence in Robotics Laboratory (AiR Lab), Department of Electrical and Computer Engineering, Aarhus University, 8000 Aarhus C, Denmark (\texttt{abdelhakim@ece.au.dk}); 
M. Mehndiratta is with NestAI, Espoo, Finland (\texttt{mohit.mehndiratta@nestai.com}); 
Y. Brodskiy is with EIVA a/s, Skanderborg, Denmark (\texttt{ybr@eiva.com}); 
B. Wehbe is with Deutsches Forschungszentrum für Künstliche Intelligenz GmbH (DFKI), Bremen, Germany (\texttt{bilal.wehbe@dfki.de}); 
E. Kayacan is with the Automatic Control Group, Department of Electrical Engineering and Information Technology, Paderborn University, Germany (\texttt{erdal.kayacan@uni-paderborn.de}).}}

\maketitle
\begin{abstract}
Inspection of underwater structures with tethered underwater vehicles is often hindered by the risk of tether entanglement. We propose REACT (real-time entanglement-aware coverage path planning for tethered underwater vehicles), a framework designed to overcome this limitation. REACT comprises a computationally efficient geometry-based tether model using the signed distance field (SDF) map for accurate, real-time simulation of taut tether configurations around arbitrary structures in 3D. This model enables an efficient online replanning strategy by enforcing a maximum tether length constraint, thereby actively preventing entanglement. By integrating REACT into a coverage path planning framework, we achieve safe and entanglement-free inspection paths, previously challenging due to tether constraints. The complete REACT framework's efficacy is validated in a pipe inspection scenario, demonstrating safe navigation and full-coverage inspection. Simulation results show that REACT achieves complete coverage while maintaining tether constraints and completing the total mission $20\%$ faster than conventional planners, despite a longer inspection time due to proactive avoidance of entanglement that eliminates extensive post-mission disentanglement. Real-world experiments confirm these benefits, where REACT completes the full mission, while the baseline planner fails due to physical tether entanglement.
\end{abstract}


\IEEEpeerreviewmaketitle



\section{Introduction}
\label{sec:introduction}

%
%
%

Operating in complex, hazardous, and otherwise inaccessible environments, \acp{ROV} have become essential for modern exploration and intervention tasks. They enable a diverse range of demanding applications, including surveying, infrastructure inspection, and deep-sea exploration \cite{amer2023unav,amer2025modelling, amer2025autonomous}, thereby expanding operational possibilities. Most \acp{ROV} are tethered to a host platform to maintain reliable communication and ensure a continuous power supply during long-duration missions. However, this tethering infrastructure introduces operational challenges related to planning and control while posing the risk of being entangled with underwater objects such as flora, fauna, or underwater structures.

Numerous \ac{CPP} algorithms are proposed in the literature for inspection-related tasks with untethered systems. In essence, they calculate approximate distance-optimal paths for a thorough inspection of 3D structures \cite{bircher2015structural,feng2024fc, amer2023visual}.  In addition, exploration path planners are employed to determine the next points of view to map unknown terrains \cite{dang2020graph}. However, these path-planning methods are restricted to untethered systems, as they do not account for possible entanglements with the surroundings. Hence, the literature still lacks tether-aware inspection planners.

\begin{figure}[t!]
	\centering	\includegraphics[width=1\linewidth]{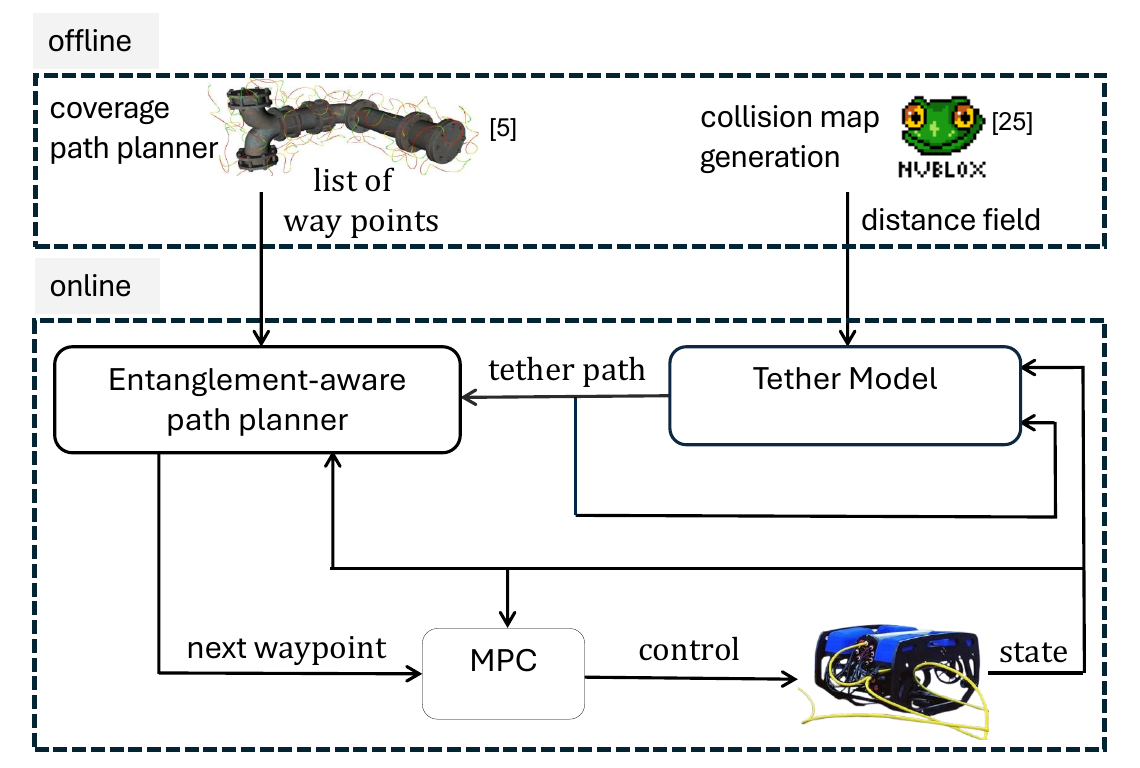}
\caption{Overview of \ac{REACT}: Offline, an \ac{SDF} map is generated from a point cloud, and an off-the-shelf \ac{CPP} \cite{feng2024fc} is used to compute a near distance-optimal waypoint sequence. Online, \ac{REACT} provides an entanglement-free next waypoint to the controller.}
    \label{fig:abstract}
\end{figure}
%

%
%
 During operation, entanglement can occur when the vehicle's movement is restricted due to interaction between the tether and objects in the environment. The tether can loop around obstacles, thereby limiting the vehicle's mobility and substantially reducing operational range in a worst-case scenario. Consequently, operators need to carry longer tethers to account for possible entanglements without a proper planner at hand.
In this work, we propose \ac{REACT} that fills this gap in underwater asset inspection. In essence, when integrated into the coverage path planning framework, \ac{REACT} renders entanglement-free paths, thus ensuring task completion without the need to carry additional tether lengths. Besides, the proposed planning framework reduces overall operational time by circumventing the post-completion detangling process, often required with traditional path-planning methods. The contributions of this work can be summarized as follows:
\begin{itemize}
\item A computationally efficient tether model that computes the entangled tether configuration.
\item An efficient online replanning method that prevents entanglement by incorporating a tether-length constraint.

\item A complete entanglement-aware inspection framework integrating \ac{CPP}, \ac{SDF}-based mapping, and \ac{MPC} to enable safe asset inspections.

\item Demonstration of the framework in simulation and real-world tests, showcasing safe and full coverage inspection.
\end{itemize}

The remainder of this paper is organized as follows: Section~\ref{sec:related_work} reviews related work. Section~\ref{sec:framework} presents the overall framework for tethered underwater inspection. Section~\ref{sec:tether_model} introduces the taut-tether model, and Section~\ref{sec:planner} describes the proposed planner. Experimental results are shown in Sections~\ref{sec:simulationexperiments} and~\ref{sec:real-world}, followed by conclusions in Section~\ref{sec:conclusion}.

\section{State of the art}
\label{sec:related_work}

The risk of entanglement with obstacles presents a serious challenge for robots with conventional path planners. To systematically address this, a taxonomy of entanglement definitions was presented in \cite{definitions}, where existing interpretations from the literature were cataloged and new definitions were introduced, paving the way for developing new path planning strategies that account for entanglement.

Initial efforts in entanglement-aware path planning focused on 2D environments and relied on offline computation. Notably, \cite{rov_mccammon} introduced the non-entangling traveling salesman problem, formulating the \ac{ROV} path planning problem as a mixed-integer program that simultaneously determines the waypoint sequence and the path’s homotopy class, guaranteeing an entanglement-free trajectory. Similarly, \cite{mechsy2017novel} addressed 2D waypoint coverage under tether constraints. While effective for predefined scenarios, these offline approaches lack adaptability in dynamic environments. More recently, \cite{peng2025spanning} proposed a spanning tree-based offline optimization for 2D coverage.


To address the need for real-time adaptability, subsequent research explored online path planning algorithms, still primarily in 2D. In \cite{kim}, a homotopy-augmented topological approach combined with graph search techniques was introduced, allowing for dynamic adjustments to the path based on environmental perception. Similarly, in \cite{withy}, a hybrid A* variant utilizing a modified tangent graph was developed. This method efficiently plans curvature-constrained paths for tethered robots subject to winding angle constraints, demonstrating guarantees and providing simulation results for online entanglement avoidance. Online cell-based decomposition strategies are also proposed to dynamically handle tether constraints and environmental changes in 2D \cite{teshnizi2014computing}.

The complexity of tether-aware path planning further increases when coordinating multiple robots. Early work in this area is presented in \cite{hert1996ties}, followed by the method proposed in \cite{zhang2019planning}. More recently, in \cite{cao2023neptune}, an efficient online path planner for multi-robot systems was presented. In this method, a homotopy-based high-level planner was integrated with trajectory optimization and smoothing techniques to generate entanglement-free paths. Similarly, \cite{teshnizi2021motion} proposed a solution to the tethered robot pair motion planning problem for tethered robots in 2D, employing a reduced visibility graph to account for tether interactions with polygonal obstacles. However, despite these advances, these approaches remain constrained to 2D environments. Moreover, while preventive paths to avoid entanglement can be planned, strategies for path planning once tether entanglement has already occurred are not provided.

Real-world applications frequently require navigation in 3D environments, such as with underwater robots. Consequently, in \cite{bhattacharya2012topological} and \cite{martinez2021optimization}, topological aspects and optimization techniques for 3D tethered navigation were explored. In \cite{petit2022tape}, a 3D exploration path planner incorporating explicit contact avoidance constraints for the tether was presented, facilitating safer navigation for single tethered robots in complex three-dimensional spaces.

The increased complexity of 3D multi-robot scenarios was addressed in \cite{hert1999motion}, where previous 2D work was extended to three dimensions. Further advances were introduced in \cite{patil2023coordinating} and \cite{cao2023path}, where path planning strategies explicitly considered the topological constraints imposed by multiple interacting tethers in 3D were proposed. While these methods advance the state of the art in multi-robot coordination, they are generally designed for offline computation and are not suited for online planning where real-time performance is needed.

In summary, existing path planners that account for tether constraints often face limitations for practical online \ac{CPP} in complex 3D settings. Many are too computationally intensive for real-time use \cite{mechsy2017novel, hert1999motion, patil2023coordinating, cao2023path}, lack integrated tether-aware CPP frameworks, or rely on simplifying assumptions such as 2D environments or basic obstacle shapes \cite{kim, withy, cao2023neptune}, hindering generalization to real-world inspection tasks. To address these limitations, \ac{REACT} is proposed, which enables real-time, entanglement-aware path planning in arbitrary 3D environments. 
\section{Overall \ac{REACT}  Framework}
\label{sec:framework}
\ac{REACT} framework, as depicted in Fig.~\ref{fig:abstract}, consists of two main components: an offline planning phase and an online execution phase. In the offline phase, the environment is provided as a point cloud, including the object to be inspected. Then, a \ac{SDF} map is generated from this point cloud using the nvblox library \cite{nvblox}. Subsequently, the extracted point cloud of the structure under inspection is processed via FC-Planner \cite{feng2024fc} to compute an optimal waypoint sequence (referred to as nominal waypoints), generating a path that renders full inspection coverage while disregarding the tether constraints.

In the online phase, tether constraints are managed by an entanglement-aware replanner that ensures that the maximum tether length is not exceeded. The proposed \ac{REACT} framework incorporates a modular online layer that can be seamlessly integrated with any existing offline coverage or path planner. By decoupling coverage planning from tether-constrained replanning, the framework transforms an otherwise highly complex and computationally intractable joint problem into two tractable subproblems: offline coverage planning and fast online tether management. This separation not only enables the use of standard off-the-shelf coverage planners, but also greatly simplifies integration with other planning frameworks.

Operating online provides further advantages: it allows the system to handle operational disturbances and model uncertainties that cannot be fully anticipated during offline planning. Moreover, it naturally supports scenarios where the map expands during operation, such as in exploration or when new obstacles are detected, enabling adaptive planning in previously unknown environments. The online management-aware planner then provides entanglement-free reference trajectories that are tracked by a \ac{MPC} controller, which translates them into optimal wrench commands for the \ac{ROV}. This ensures that the vehicle follows the re-planned paths while respecting dynamics and actuation limits.

\section{Tether Modeling}
\label{sec:tether_model}

 In this section, we describe a computationally efficient, geometry-based tether model for tethered underwater vehicles. This model predicts the tether path, specifically the positions of each node along the tether's length, based on the \ac{ROV}'s trajectory, and assumes a geometry-based constraint where the tether remains taut and fully stretched at all times. The main idea of the proposed tether model is inspired by the shortcutting algorithm of the ropeRRT path planner \cite{roperrt}, which simplifies sampled trajectories similar to a rope tightening around an obstacle.

\subsection{Tether model description}

To efficiently compute the tether configuration around obstacles, the continuous tether is discretized into a finite set of nodes, separated by a fixed resolution $\delta$. Each node represents a point along the tether and collectively approximates the overall 3D shape of the tether. This discretization enables computationally efficient collision checking and shortcutting.

\begin{figure*}[t]
    \centering
    \includegraphics[width=0.97\linewidth]{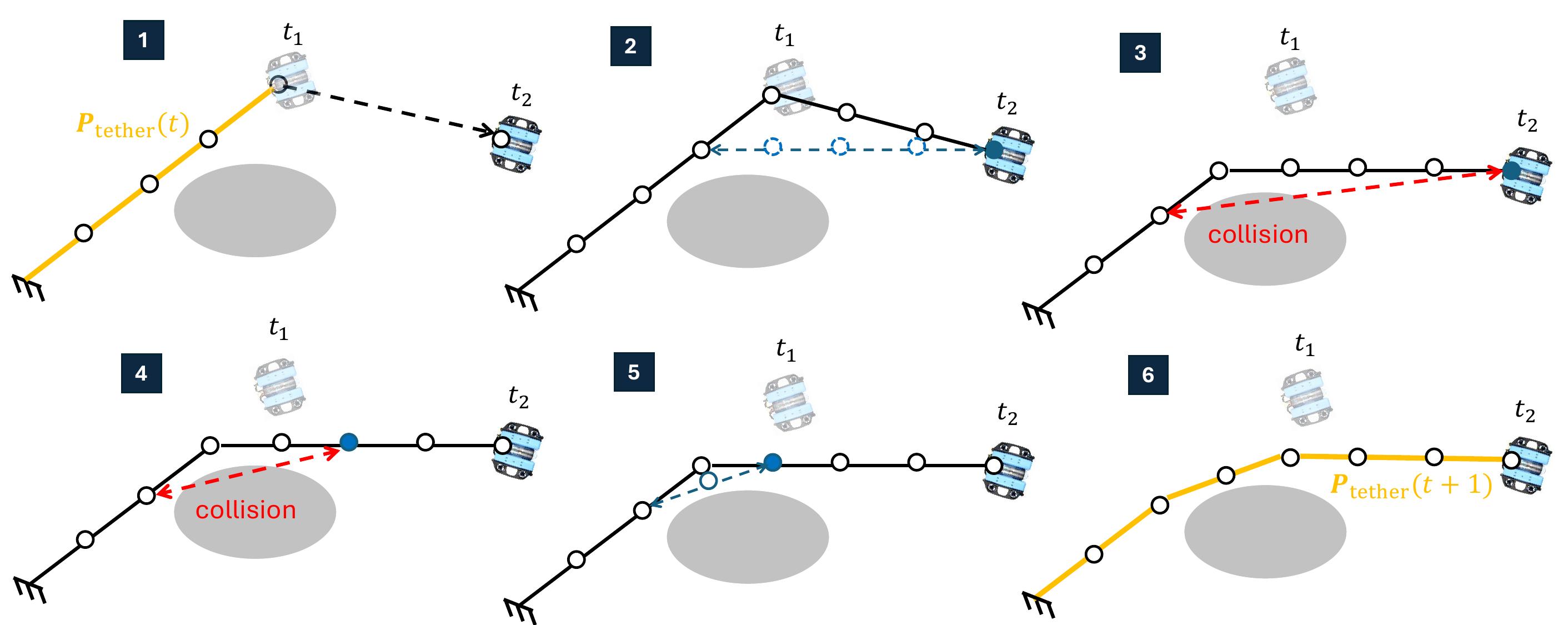}
    \caption{Tether shortcutting during \ac{ROV} motion from \( t_1 \) to \( t_2 \). (1) Initial tether with new \ac{ROV} position \( \mathbf{p}_{\text{rov}} \) appended; (2) Successful shortcut from the end node; (3) Collision encountered when attempting further shortcutting, skipping to the next node; (4) Another collision detected from the new node; (5) Successful shortcut from a subsequent node; (6) Final tether configuration (yellow) after applying all feasible shortcuts.}
    \label{fig:tether}
\end{figure*}

Let the tether path at time \( t \) be denoted by \( \mathbf{P}_{\mathrm{tether}}(t) = \{ \mathbf{p}_i(t) \}_{i=1}^{n} \), where each node \( \mathbf{p}_i(t) \in \mathbb{R}^3 \) represents the position of the \( i \)-th node in 3D space at time \( t \), and \( n \) is the total number of nodes in the tether path. The \ac{ROV} position at time \( t \), denoted by \( \mathbf{p}_{\mathrm{rov}}(t) \in \mathbb{R}^3 \), is appended at the end \( \mathbf{p}_{n+1}(t) \).  The proposed tether model then iteratively computes the equivalent taut-tether path through sequential shortcutting operation. Fig.~\ref{fig:tether} illustrates an example of the shortcut operation. Starting at the last node ($\mathbf{p}_j(t) = \mathbf{p}_{n}(t)$), the algorithm attempts to shortcut the path segment between each pair of nodes \( (\mathbf{p}_i(t), \mathbf{p}_j(t)) \) where \( i > j \).
If the line of sight between \( \mathbf{p}_i(t) \) and \( \mathbf{p}_j(t) \) is collision-free, as determined via an \ac{SDF} map (\( \mathcal{M}_{sdf} \)), the intermediate nodes are replaced with a straight segment sampled at known resolution \( \delta \). Conversely, if the line of sight encounters a collision, $j$ shifts to the preceding node, and the collision-free line of sight is rechecked. This process is repeated until a collision-free line of sight is found, and the intermediate nodes are then replaced.  Finally, after the shortcutting step, a pulling operation is applied to each node, moving it incrementally toward the tether endpoint \( \mathbf{p}_{n+1}(t) \). This ensures that nodes are not left stuck in the cavities of non-convex obstacles. The shortcutting and pulling operations are applied iteratively until convergence, resulting in a taut and collision-free tether path \( \mathbf{P}_{\mathrm{tether}}(t+1) \). The full procedure is described in ~\ref{alg:tether_optimization}.

\begin{remark}
Sparse environment reconstructions may create gaps in the \ac{SDF}, causing obstacles to appear partially free. This can lead to unrealistic tether shortcutting through obstacles unless sufficient voxel resolution is applied.
\end{remark}

\begin{remark}
The tether discretization resolution \( \delta \) should be chosen consistently with the \ac{SDF} voxel resolution to ensure reliable collision detection.
\end{remark}



\begin{algorithm}[t]
\LinesNotNumbered  

\SetKwInOut{Input}{Input}
\SetKwInOut{Output}{Return}
\Input{
$\mathbf{p}_{\mathrm{rov}}(t)$, 
$\mathbf{P}_{\mathrm{tether}}(t)$, 
$\mathcal{M}_{sdf}$, 
$\delta$
}
\Output{$\mathbf{P}_{\mathrm{tether}}(t+1)$}
\BlankLine

appendPath($\mathbf{P}_{\mathrm{tether}}(t)$, $\mathbf{p}_{\mathrm{rov}}(t)$)\;

\For{$i \gets \mathrm{len}(\mathbf{P}_{\mathrm{tether}}(t)) - 1$ \textbf{to} $0$}{
    \For{$j \gets i - 1$ \textbf{to} $0$}{
        \If{$\mathrm{checkShortcut}$($\mathbf{P}_{\mathrm{tether}}(t), i, j$)}{
            replaceNodes($\mathbf{P}_{\mathrm{tether}}(t)$, $i$, $j$, $\delta$)\;
        }
        \Else{
            \If{not $\mathrm{checkLineOfSight}$
            ($\mathcal{M}_{sdf}$, $\mathbf{P}_{\mathrm{tether}}(t)$)}{
                \textbf{break}\;
            }
            \If{$\mathrm{isInCollision}$($\mathcal{M}_{sdf}$, $\mathbf{P}_{\mathrm{tether}}(t)[j]$)}{
                pullNode($\mathbf{P}_{\mathrm{tether}}(t)[j]$, $\mathbf{P}_{\mathrm{tether}}(t).end()$, $\delta$)\;
            }
        }
    }
}
\Return{$\mathbf{P}_{\mathrm{tether}}(t+1)$}\;
\caption{Taut-tether model}
\label{alg:tether_optimization}
\end{algorithm}


\section{Entanglement-Aware Path Planner}
\label{sec:planner}

Next, we present the entanglement-aware path planner, a local planner designed to address the real-time entanglement avoidance problem for tethered underwater vehicles. The planner continuously monitors the tether configuration and dynamically adjusts the vehicle’s target to prevent the tether length from exceeding a specified maximum allowable length, \( L_{\max} \), while ensuring safe navigation toward a series of predefined reference waypoints.

At each time step \( t \), the planner receives the current estimated tether path, \(\mathbf{P}_{\mathrm{tether}}(t)\), from the tether model. It also maintains an ordered list of target waypoints, \(\mathbf{W} = \{\mathbf{p}_{\text{waypoint}}(k)\}_{k=1}^m\), where each \(\mathbf{p}_{\text{waypoint}}(k)\) specifies a 3D position the ROV must reach sequentially. The current position of the ROV, \(\mathbf{p}_{\text{rov}}(t)\), and the current waypoint index, \(k\), are also maintained as part of the planner’s state. The current tether length \(L_{\text{tether}}(t)\) is computed from the tether path and compared against the maximum allowable tether length \(L_{\max}\) to determine whether replanning is needed.

The planner operates in two main modes: normal mode and recovery mode. In normal mode, when the tether length is within the allowable limit, the planner sets the ROV’s target position directly to the current waypoint. It continuously monitors the ROV’s position and, upon detecting that the waypoint has been reached, advances the waypoint index to the next target in the sequence. This mode prioritizes nominal waypoint following behavior.

If the tether length exceeds the limit, the planner switches to recovery mode to avoid entanglement. In this mode, a recovery path is generated by searching for an alternative safe path toward the current waypoint that respects the tether length constraint. The ROV’s target position, $\mathbf{p}_{\text{target}}$,  is then updated to incrementally follow this recovery path. The planner remains in recovery mode until the end of the recovery path is reached, at which point it switches back to normal mode to continue waypoint tracking.

This dual-mode approach ensures that the vehicle maintains a safe tether configuration while effectively progressing through its mission waypoints. By continuously switching between nominal waypoint tracking and recovery behaviors in response to tether length measurements, the planner enables real-time, entanglement-aware path planning. The overall planning framework is summarized in Algorithm \ref{alg:main_loop}.

\begin{algorithm}[t]
\LinesNotNumbered  
\SetKwInOut{Input}{Input}
\SetKwInOut{Output}{Output}
\Input{
$\mathbf{W}$, 
$L_{\max}$, 
$\mathbf{P}_{\mathrm{tether}}(t)$, 
$\mathbf{p}_{\text{rov}}(t)$
}
\Output{$\mathbf{p}_{\text{target}}$}
\BlankLine
$mode \gets \mathrm{NORMAL}$\;
$\mathbf{P}_{\text{recovery}} \gets \emptyset$\;
$k \gets 0$\;
\BlankLine
$L_{\text{tether}}(t) \gets \text{computeLength}\bigl(\mathbf{P}_{\mathrm{tether}}(t)\bigr)$\;
\BlankLine
\If{$mode = \mathrm{NORMAL}$}{
    \If{$L_{\mathrm{tether}}(t) > L_{\max}$}{
        $mode \gets \mathrm{RECOVERY}$\;
        $\mathbf{P}_{\text{recovery}} \gets \text{disEntanglementSearch}\bigl(\mathbf{P}_{\mathrm{tether}}(t), \mathbf{W}[k]\bigr)$\;
        $\mathbf{p}_{\text{target}} \gets \text{followPath}\bigl(\mathbf{P}_{\text{recovery}}, \mathbf{p}_{\text{rov}}(t)\bigr)$\;
    }
    \Else{
        $\mathbf{p}_{\text{target}} \gets \mathbf{W}[k]$\;
        \If{$\mathrm{reachedWaypoint}\bigl(\mathbf{p}_{\text{rov}}(t), \mathbf{W}[k]\bigr)$}{
            $k \gets k + 1$\;
        }
    }
}
\ElseIf{$mode = \mathrm{RECOVERY}$}{
    $\mathbf{p}_{\text{target}} \gets \text{followPath}\bigl(\mathbf{P}_{\text{recovery}}, \mathbf{p}_{\text{rov}}(t)\bigr)$\;
    \If{$\mathrm{reachedEndOfPath}\bigl(\mathbf{p}_{\text{rov}}(t), \mathbf{P}_{\text{safe}}\bigr)$}{
        $mode \gets \mathrm{NORMAL}$\;
    }
}
\BlankLine
\Return{$\mathbf{p}_{\mathrm{target}}$}\;
\caption{Entanglement-aware path planner}
\label{alg:main_loop}
\end{algorithm}

\subsection{Disentanglement path search}

To implement the disentanglement behavior described above, the recovery path search algorithm (Algorithm~\ref{alg:search_alternative}) is employed. This algorithm takes as input the current tether path \(\mathbf{P}_{\mathrm{tether}}(t)\), the current target waypoint \(\mathbf{W}[k]\), and the maximum allowable tether length \(L_{\max}\). Its goal is to generate a recovery path \(\mathbf{P}_{\mathrm{recovery}}\) that guides the \ac{ROV} safely toward the waypoint while respecting the tether length constraint.

The algorithm performs a backward search along the tether path, starting from the \ac{ROV}’s current position at the tether’s end and progressing toward toward the tether starting node ($\mathbf{p}_0$). At each iteration, a candidate pivot point \(\mathbf{p}_{\mathrm{pivot}}\) on the tether is selected as a potential point from which the \ac{ROV} can attempt an alternative route toward the target waypoint.

For each pivot, two path segments are constructed. The first segment, \(\mathbf{P}_{\mathrm{r}1}\), corresponds to the reversed portion of the tether from the pivot \(\mathbf{p}_{\mathrm{pivot}}\) to the \ac{ROV}’s current position, representing the retraced section of the tether. The second segment, \(\mathbf{P}_{\mathrm{s}(n - i)}\), is a newly planned shortest path connecting the pivot to the target waypoint \(\mathbf{W}[k]\), which can be computed using sampling-based methods such as RRT*. These segments are then concatenated with the tether path from $\mathbf{p}_0$ up to the pivot, forming an augmented candidate path \(\mathbf{P}_{\mathrm{aug}}\), which is used to simulate the tether configuration and estimate its length \(L_{\mathrm{tether}}\) at the next time step. The total tether length \(L_{\mathrm{tether}}\) is then compared against the maximum allowable length \(L_{\max}\). If the length constraint is satisfied, the recovery path \(\mathbf{P}_{\mathrm{recovery}} = \mathbf{P}_{\mathrm{r}1} \cup \mathbf{P}_{\mathrm{s}(n - i)}\) is considered feasible.

\begin{figure*}[t]
    \centering
    \includegraphics[width=\textwidth]{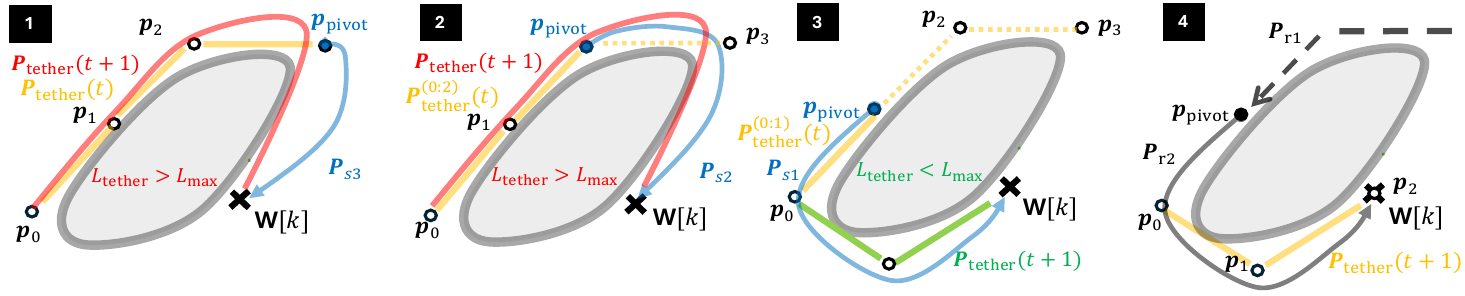}
    \caption{Disentanglement path search process. At time \( t \), the tether configuration is \( \mathbf{P}_{\text{tether}}(t) = \{\mathbf{p}_0, \mathbf{p}_1, \mathbf{p}_2, \mathbf{p}_3\} \). 
    (1) The shortest path \( \mathbf{P}_{s3} \) from \( \mathbf{p}_3 \) to the waypoint \( \mathbf{W}[k] \) is appended to the tether path, forming the augmented path \( \mathbf{P}_{\text{tether}}^{(0:3)} \cup \mathbf{P}_{s3} \), which is passed to the tether model to compute \( \mathbf{P}_{\text{tether}}(t+1) \). The resulting tether length exceeds \( L_{\max} \), indicating a violation. 
    (2) The pivot shifts to \( \mathbf{p}_2 \), and the augmented path \( \mathbf{P}_{\text{tether}}^{(0:2)} \cup \mathbf{P}_{s2} \) is evaluated, but still violates the constraint. 
    (3) With pivot at \( \mathbf{p}_1 \), the augmented path \( \mathbf{P}_{\text{tether}}^{(0:1)} \cup \mathbf{P}_{s1} \) yields a feasible configuration \( \mathbf{P}_{\text{tether}}(t+1) \) with \( L_{\text{tether}} \leq L_{\max} \). 
    (4) The recovery path \( \mathbf{P}_{\text{recovery}} \)  is thus executed, consisting of \( \mathbf{P}_{\text{r}1} \), tracing back from \( \mathbf{p}_3 \) to \( \mathbf{p}_1 \), and \( \mathbf{P}_{\text{r}2} = \mathbf{P}_{s1} \), leading to the waypoint \( \mathbf{W}[k] \).}
    \label{fig:planner_search}
\end{figure*}

If no feasible path is found after examining all possible pivot points, the algorithm defaults to a fallback strategy where the tether length constraint is treated as a soft limit. In this case, the recovery path consists solely of the shortest path from the \ac{ROV}’s current position directly to the target waypoint, temporarily disregarding tether length constraints. The resulting recovery path provides a safe and viable trajectory for the \ac{ROV} to follow during disentanglement operations by combining the reversed tether segment with the newly planned path to the waypoint, as shown in Fig. ~\ref{fig:planner_search}.

\begin{algorithm}[t]
\LinesNotNumbered
\SetKwInOut{Input}{Input}
\SetKwInOut{Output}{Output}
\Input{
$ \mathbf{P}_{\text{tether}}(t), \mathbf{W}[k], L_{\max} $
}
\Output{
Recovery path $ \mathbf{P}_{\text{recovery}} $
}
\BlankLine
$found\_feasible \gets \textbf{False}$\;
$n \gets \text{len}\bigl(\mathbf{P}_{\text{tether}}(t)\bigr)$\;
\BlankLine
\For{$i \gets n-1$ \textnormal{ downto } $0$}{
    $\mathbf{p}_{\text{pivot}} \gets \mathbf{P}_{\text{tether}}(t)[i]$\;
    $\mathbf{P}_{\text{r}1} \gets \text{reverseSegment}\bigl(i, n-1, \mathbf{P}_{\text{tether}}(t)\bigr)$\;
    $\mathbf{P}_{\text{s}(n - i)} \gets \text{planShortestPath}\bigl(\mathbf{p}_{\text{pivot}}, \mathbf{W}[k]\bigr)$\;
    \BlankLine
    $\mathbf{P}_{\text{aug}} \gets \text{concatenate}\bigl(\mathbf{P}_{\text{tether}}(t)[0:i], \mathbf{P}_{\text{s}(n - i)}\bigr)$\;
    $\mathbf{P}_{\text{tether}}(t+1) \gets \text{tetherModel}\bigl(\mathbf{P}_{\text{aug}}\bigr)$\;
    $L_{\text{tether}} \gets \text{computeLength}\bigl(\mathbf{P}_{\text{tether}}(t+1)\bigr)$\;
    \BlankLine
    \If{$L_{\text{tether}} \leq L_{\max}$}{
        $\mathbf{P}_{\text{r}2} \gets \mathbf{P}_{\text{s}(n - i)}$\;
        $found\_feasible \gets \textbf{True}$\;
        \textbf{break}\;
    }
}
\BlankLine
\If{\textbf{not} $found\_feasible$}{
    $\mathbf{p}_{\text{pivot}} \gets \mathbf{P}_{\text{tether}}(t)[-1]$\;
    $\mathbf{P}_{\text{r}1} \gets \emptyset$\;
    $\mathbf{P}_{\text{r}2} \gets \text{planShortestPath}\bigl(\mathbf{p}_{\text{pivot}}, \mathbf{W}[k]\bigr)$\;
}
\BlankLine
$\mathbf{P}_{\text{recovery}} \gets \mathbf{P}_{\text{r}1} \cup \mathbf{P}_{\text{r}2}$\;
\Return{$\mathbf{P}_{\mathrm{recovery}}$}\;
\caption{Disentanglement path search}
\label{alg:search_alternative}
\end{algorithm}

\subsection{Recovery path refinement}

Once the recovery path \(\mathbf{P}_{\text{recovery}}\) is generated through the disentanglement search, it undergoes a refinement process to generate a safe path \(\mathbf{P}_{\text{safe}}\) with obstacle clearance \(\delta_{obst}\) and improved smoothness prior to execution. This refinement is performed iteratively and consists of three consecutive operations: centroid offset, stochastic local perturbation, and polynomial smoothing. The resulting safe trajectory \(\mathbf{P}_{\text{safe}}\) is validated to ensure collision-free execution.

The centroid offsetting step is designed to increase clearance from obstacles that may lie near the geometric center of the path. The geometric centroid of the path is calculated as the mean of all points, \(\mathbf{c} = \frac{1}{N}\sum_{i=1}^{N} \mathbf{p}_i\), where \(N\) denotes the number of points in \(\mathbf{P}_{\text{recovery}}\) and \(\mathbf{p}_i\) represents the \(i\)-th point. Each point is then displaced along the normalized vector that points from the centroid to the point, \(\mathbf{d}_i = (\mathbf{p}_i - \mathbf{c}) / \|\mathbf{p}_i - \mathbf{c}\|\), producing the updated position \(\mathbf{p}_i' = \mathbf{p}_i + \delta_{obst} \mathbf{d}_i\). This operation effectively increases the clearance from the obstacles.

Following centroid offsetting, the path is further refined using stochastic local perturbations. Each intermediate point along the path (excluding the endpoints) is locally adjusted by sampling random directions uniformly on the surface of a sphere and displacing the point along the sampled direction by a small distance. The resulting candidate points are evaluated for collisions and only those that are collision-free are accepted. Multiple trials are allowed for each point to maximize the likelihood of finding a feasible local adjustment. This perturbation mechanism adjusts the path points to locally increase clearance from nearby obstacles, ensuring the trajectory remains safely away from potential collisions while preserving its overall structure. The final stage of refinement involves smoothing the path using a third-order polynomial fitting. The path is divided into overlapping segments and for each segment separate cubic polynomials are fitted to the x, y, and z coordinates of the points using a least squares approximation. Each point in the segment is then replaced by the corresponding evaluation of the polynomial functions, producing a smoother trajectory. The overlapped segments ensure continuity between the polynomial pieces and prevent abrupt changes in curvature, resulting in a dynamically feasible path suitable for the \ac{ROV}. After completion of these operations, the refined path is subjected to collision verification. If any collisions are detected, the refinement process is repeated iteratively until a collision-free trajectory is obtained.


\section{Simulations Experiments}
\label{sec:simulationexperiments}

This section presents the simulation results for the proposed \ac{REACT} method. The entire framework is implemented in C++ to ensure computational efficiency and is integrated with ROS to facilitate modularity and deployment on real-world robots. Furthermore, the \ac{OMPL} library \cite{ompl} is utilized to compute the shortest paths.

\subsection{Simulation experimental setup}
The path planner is implemented for a BlueROV2 underwater robot. An \ac{MPC} approach accounts for model constraints and provides the optimal control input $\mathbf{u}^{ref} \in \mathbb{R}^4$, where $\mathbf{u}^{ref} = [F_x, F_y, F_z, M_z]^T$ represents the forces in the $X$, $Y$, and $Z$ directions, and the moment about the $Z$-axis. The goal is to follow the desired reference trajectory \(\mathbf{x}^{\text{ref}} \in \mathbb{R}^4\), where \(\mathbf{x}^{\text{ref}} = [x^{\text{ref}}, y^{\text{ref}}, z^{\text{ref}}, \psi^{\text{ref}}]^T\) contains the reference position \((x^{\text{ref}}, y^{\text{ref}}, z^{\text{ref}})\) and the reference yaw angle \(\psi^{\text{ref}}\). For further details about the controller and model, the reader is referred to \cite{amergp}. The entanglement-aware path planner provides $\mathbf{x}^{ref}$ in real-time for the MPC.

We perform a comparative analysis of the proposed \ac{REACT} method and a baseline conventional \ac{CPP} (FC-Planner \cite{feng2024fc}), which does not explicitly handle entanglement. 
The simulation setup uses a 1/10-scale pipe structure based on the model described in \cite{feng2024fc}. The simulated onboard camera has a $70$-degree field of view. The tether constraint is implemented by specifying a maximum allowable tether length of \(L_{\text{max}} = 10\)m. Note that the resulting planned trajectories are geometrically identical to those in the full-scale environment, simply scaled down proportionally. The primary motivation for using the scaled model is to accelerate simulation runs while preserving the validity of the results. Coverage is computed in a manner similar to \cite{amer2023visual}. At each time step, the position and orientation of the camera are used to determine which triangles in the environment mesh are visible. A triangle is marked as visible if its centroid is within the inspection range, its surface normal faces the camera, and its projection lies within the camera's field of view. Over time, the set of all uniquely observed triangles accumulates. The coverage at time $t$ is defined as the ratio of the number of unique visible triangles up to time $t$ to the total number of triangles in the environment.

\subsection{Simulations results}

\begin{table}[b]
    \centering
    \caption{Coverage performance comparison between \ac{REACT} and \ac{CPP} baseline showing inspection time, recovery time, total mission duration, and final coverage. }
    \label{tab:performance_metrics}
    {\large
    \resizebox{1.0\columnwidth}{!}{%
    \begin{tabular}{|l|c|c|c|c|}
        \hline
        \textbf{Planner} & \textbf{Inspection time (s)} & \textbf{Recovery time (s)} & \textbf{Total time (s)} & \textbf{Coverage (\%)} \\
        \hline
        \ac{REACT} & 546 & 134 & \textbf{680} & \textbf{99.91} \\
        CPP & 429 & 426 & 855 & 99.82 \\
        \hline
    \end{tabular}%
    }
    }
    \vspace{0.5em}
\end{table}

\begin{table}[b]
    \centering
    \caption{Comparison of tether constraint compliance.}
    \label{tab:tether_metrics}
    {\Large
    \resizebox{1.0\columnwidth}{!}{%
    \begin{tabular}{|l|c|c|}
        \hline
        \textbf{Planner} & \textbf{Maximum tether length (m)} & \textbf{Duration of length exceedance (s)} \\
        \hline
        CPP & 31.16 & 327.37 \\
        \ac{REACT} & \textbf{10.52} & \textbf{10.36} \\
        \hline
    \end{tabular}%
    }
    }
    \vspace{0.5em}
\end{table}

\begin{figure}[b]
    \centering
    \begin{subfigure}[b]{0.48\linewidth}
        \centering
        \includegraphics[width=\linewidth]{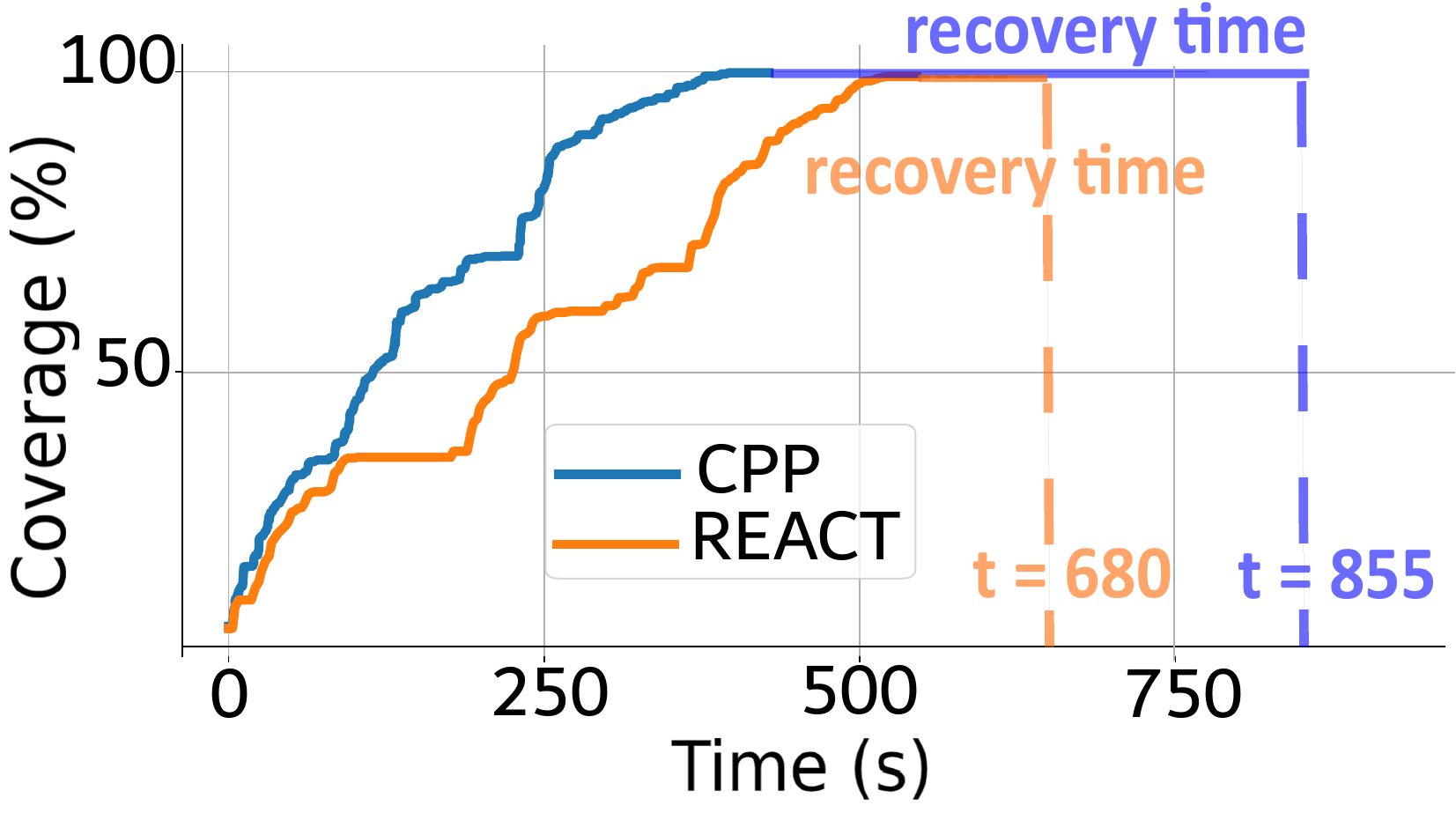}
        \caption{Coverage vs. time.}
        \label{fig:coverage_vs_time}
    \end{subfigure}
    \hfill 
    \begin{subfigure}[b]{0.48\linewidth}
        \centering
        \includegraphics[width=\linewidth]{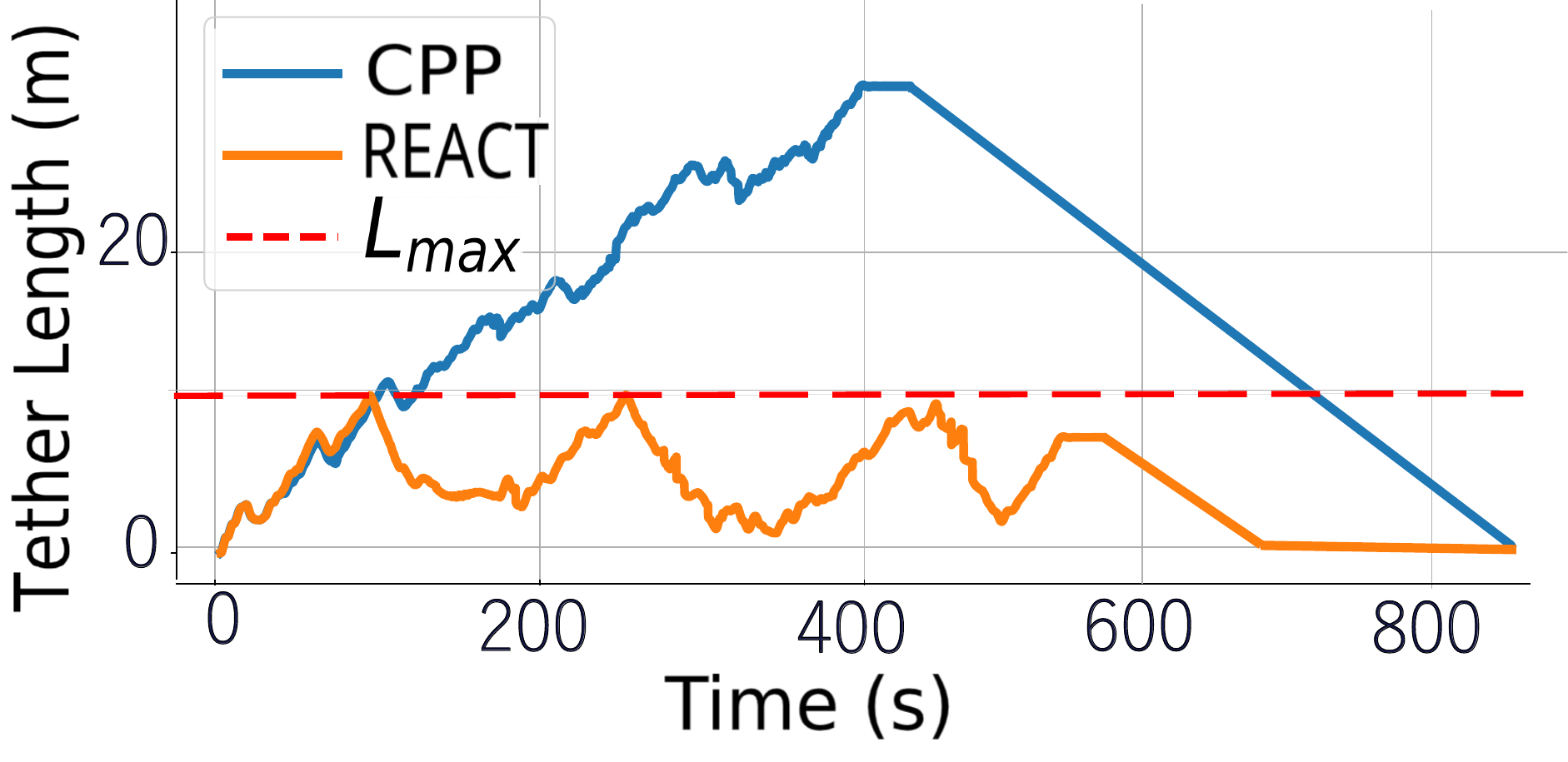}
        \caption{Tether Length vs. time.}
        \label{fig:tether_vs_time}
    \end{subfigure}
    \caption{Comparison of coverage and tether length.}
    \label{fig:coverage_tether_sidebyside}
\end{figure}

\begin{figure}[t]
    \centering
    \begin{subfigure}[b]{0.48\linewidth}
        \centering
        \includegraphics[width=\linewidth]{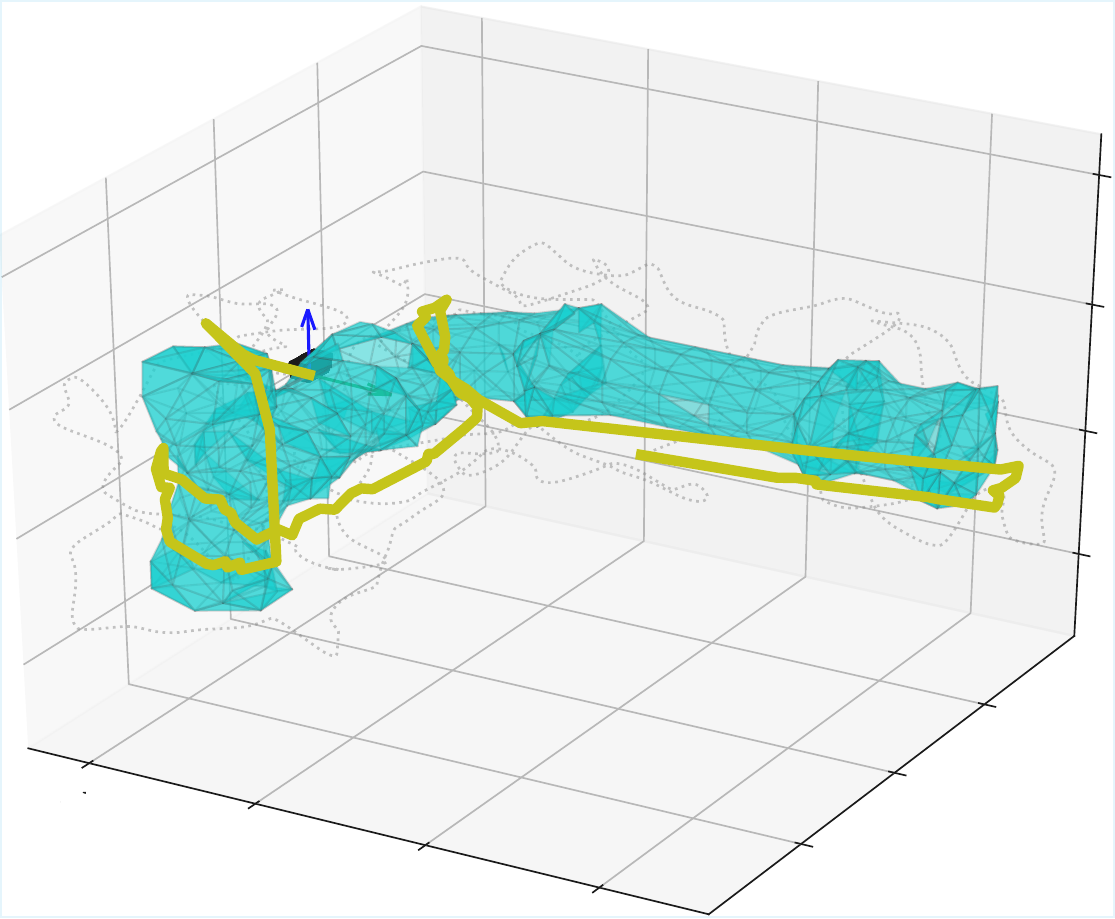}
        \caption{\ac{CPP} final tether path.}
        \label{fig:3d_cpp}
    \end{subfigure}
    \hfill
    \begin{subfigure}[b]{0.48\linewidth}
        \centering
        \includegraphics[width=\linewidth]{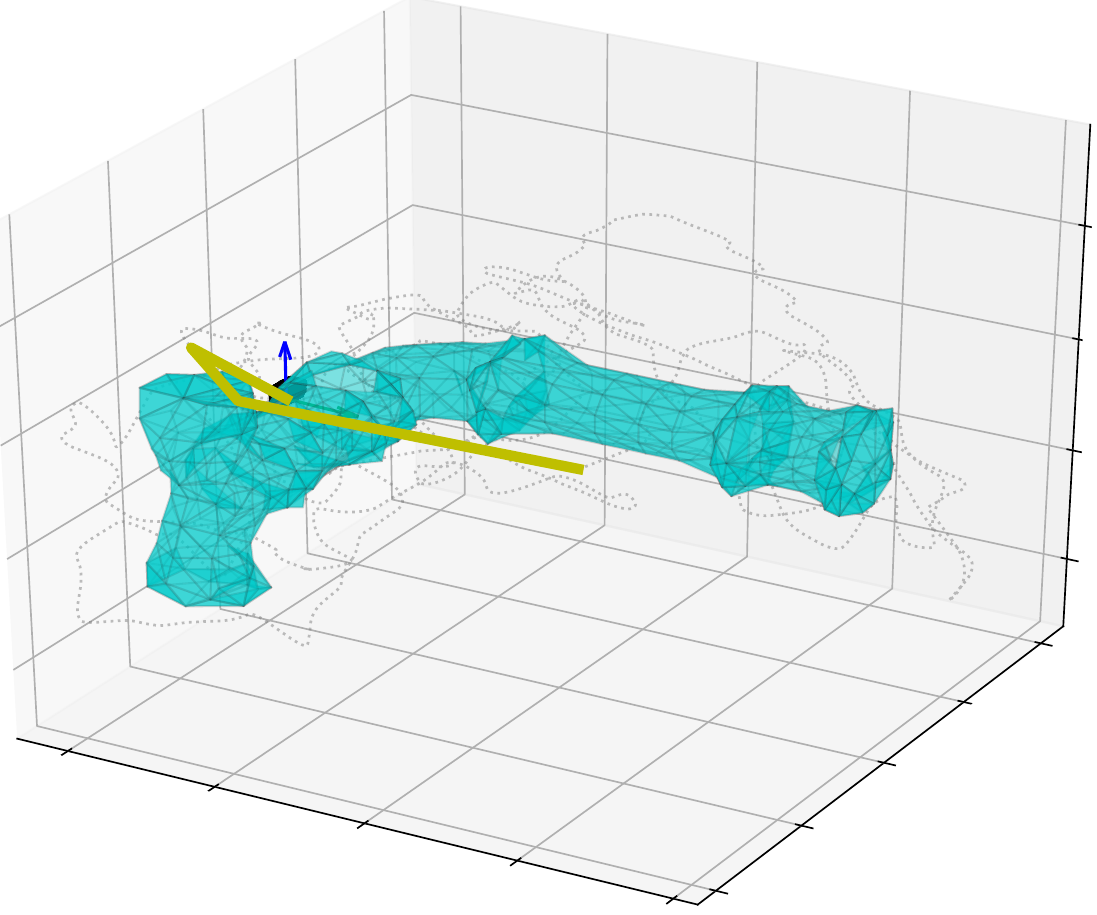}
        \caption{\ac{REACT} final tether path.}
        \label{fig:3d_oea}
    \end{subfigure}
    \caption{3D views of final trajectories: (a) CPP results in entangled tether path, (b) \ac{REACT} yields a non-entangled tether path, reflecting effective entanglement avoidance.}
    \label{fig:3Dplots}
\end{figure}

The performance metrics for both planners are summarized in Table~\ref{tab:performance_metrics} and Table~\ref{tab:tether_metrics}, while Figs.~\ref{fig:coverage_tether_sidebyside} and~\ref{fig:3Dplots} visualize the performance of the planners in terms of environmental coverage, tether length behavior, and final trajectory configurations. The evaluation focuses on comparing mission efficiency, constraint compliance, and safety aspects between the entanglement-aware \ac{REACT} method and the conventional \ac{CPP}..

The results present two phases: the inspection phase and the recovery phase, where the recovery time represents the estimated duration required to return to the starting position after complete inspection while disentangling the entire tether.
The results highlight distinct trade-offs between the two planning strategies. As shown in Table~\ref{tab:performance_metrics}, the \ac{CPP} method achieves a shorter inspection time ($429$s vs. $546$s) due to its straightforward path execution without rerouting for disentanglement. However, focusing solely on inspection speed overlooks the critical aspect of tether management in constrained environments. The \ac{CPP} exhibits a significantly longer total mission time ($855$s vs. $680$s) because extensive disentanglement is required after inspection completion.

\ac{REACT} demonstrates multiple instances of entanglement detection and resolution, as evidenced by the peaks in the tether length curve in Fig.~\ref{fig:tether_vs_time} and the corresponding flat regions in the coverage curve in Fig.~\ref{fig:coverage_vs_time}, where the \ac{ROV} inspection progress stops to reroute and find paths without entanglement. This reactive replanning behavior directly results in a longer inspection time, as the system prioritizes tether safety over raw speed. In particular, the limit on the length of the tether is rarely exceeded by \ac{REACT}, as shown in Table~\ref{tab:tether_metrics} and Fig.~\ref{fig:tether_vs_time}. Conversely, the \ac{CPP} method severely violates this constraint, reaching $31.16$m for extended periods ($327.37$s), risking unrecoverable tether entanglement. The 3D trajectory visualizations in Fig.~\ref{fig:3Dplots} further illustrate this difference, showing the entangled tether geometry resulting from \ac{CPP} versus the non-entangled configuration achieved by \ac{REACT}. Therefore, \ac{REACT} demonstrates superior performance for safe tethered \ac{ROV} inspection.
\section{Real-world Experiments}
\label{sec:real-world}

\begin{figure}[t]
    \centering
    \includegraphics[width=0.75\linewidth]{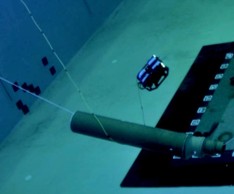}
    \caption{
    During the experiment, the ROV became entangled with the pipe, illustrating a typical \ac{CPP} entanglement.}
    \label{fig:pipe}
\end{figure}

 Real-world experiments were conducted at $23$m$\times19$m$\times8$m test basin equipped with a 12-camera Qualisys motion capture system. The experimental platform is a BlueROV2, configured identically to the simulation setup. The test environment consists of a pipe structure with a diameter of $0.7$m and a length of $2.5$m, as shown in Fig.~\ref{fig:pipe}. Two planning modes, identical to those evaluated in simulation, are tested. The first is \ac{REACT}, which represents the full system incorporating \ac{REACT}. The second is the \ac{CPP} baseline without \ac{REACT}. In both modes, a helical trajectory is given as a reference.

\subsection{Real-world experimental results}

\begin{figure}[t]
    \centering
    \begin{subfigure}[b]{0.48\linewidth}
        \centering
        \includegraphics[width=\linewidth]{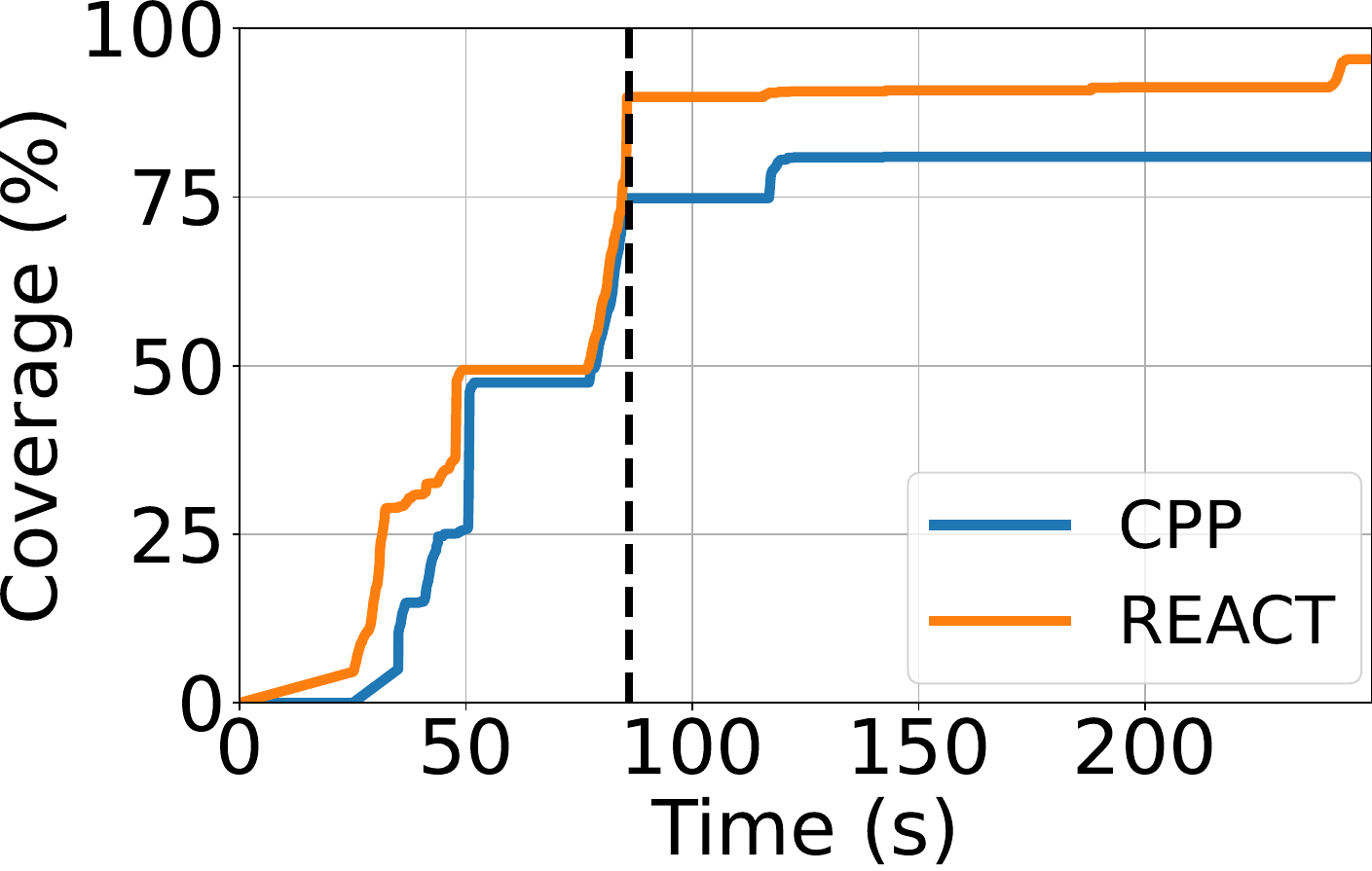}
        \caption{Coverage vs. time for both planners.}
        \label{fig:traj_noreact}
    \end{subfigure}
    \hfill
    \begin{subfigure}[b]{0.48\linewidth}
        \centering
        \includegraphics[width=\linewidth]{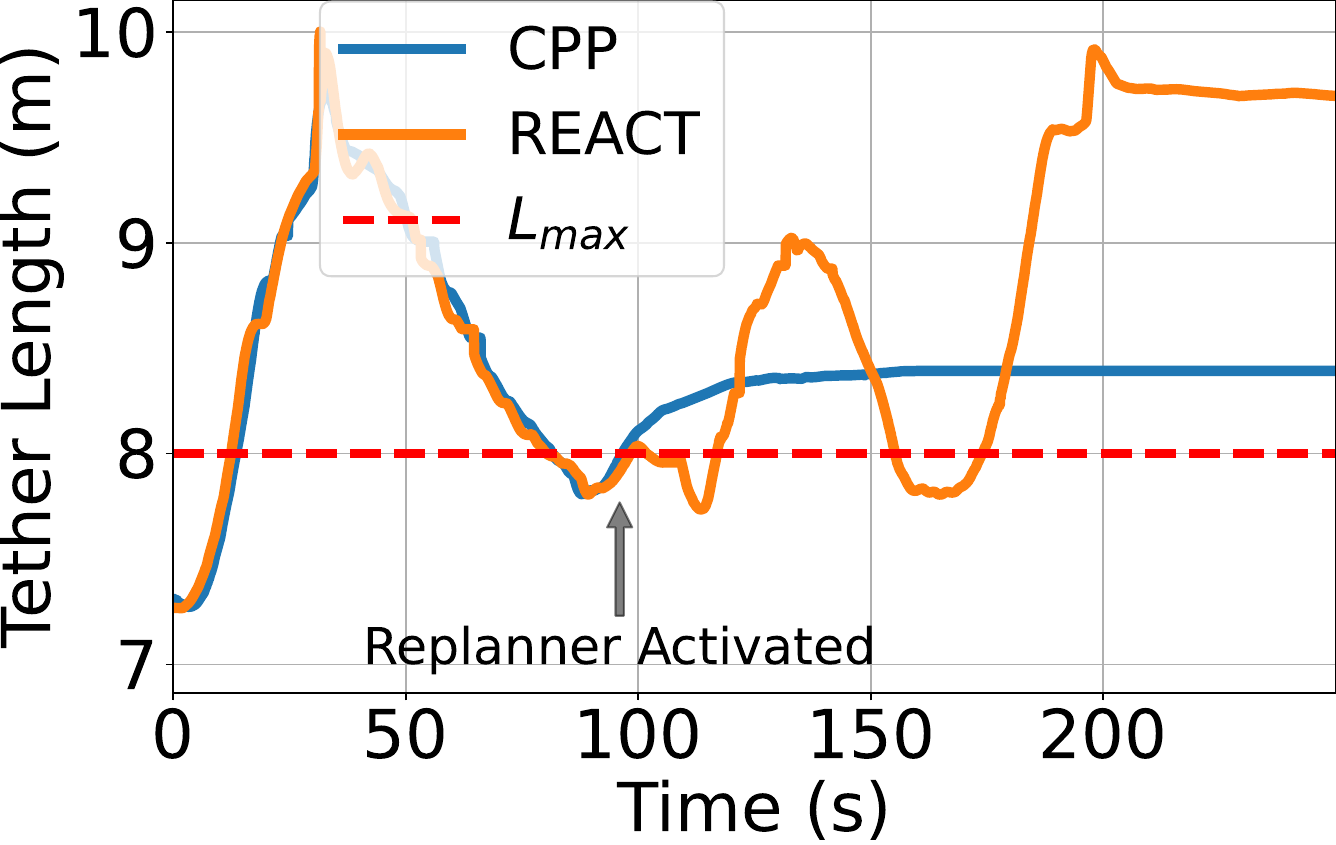}
        \caption{Tether length vs. time for both planners.}
        \label{fig:traj_react}
    \end{subfigure}
    \caption{Real-world ROV trajectories. Without \ac{REACT}, the mission fails due to tether entanglement. With \ac{REACT}, the ROV completes the mission successfully. At $t = 96$s, the replanner is activated, allowing \ac{REACT} to disentangle the \ac{ROV} and continue the inspection. In contrast, at $t = 124$s, the \ac{ROV} using the conventional \ac{CPP} becomes stuck.}
    \label{fig:realworld_trajectory}
\end{figure}

Real-world tests illustrate the need of \ac{REACT} compared to classic \ac{CPP} approach. Figure~\ref{fig:realworld_trajectory} shows the mission trajectories for both methods.
As such, the baseline \ac{CPP} planner failed to complete the inspection task. At $t=124$ s, the tether entangles the pipe structure, stopping the \ac{ROV} from further progress. Hence, the overall coverage achieved is only $80.9\%$ (Table~\ref{tab:realworld_coverage}).

The proposed \ac{REACT} planner completed the full mission, achieving $95.6\%$ coverage. The planner adapted to tether constraints throughout the mission. Unlike the simulation results, the soft tether length limit of the $8$m was exceeded in real-world tests. As such, when no entanglement-free paths to the goal are available, \ac{REACT} treats the length constraints as a soft limit rather than a hard limitation. However, when entanglement is detected and there is an entanglement-free path around the pipe, \ac{REACT} finds a path that detangles the \ac{ROV} from the structure. This capability is demonstrated at $t = 96$s when \ac{REACT} detected an imminent tether entanglement. The planner computed a new path that safely untangled the \ac{ROV} from the pipe structure, allowing inspection to continue. The system only exceeded the soft tether limit when necessary for mission completion while prioritizing entanglement avoidance whenever possible.

The baseline \ac{CPP} planner, lacking entanglement-awareness, became permanently stuck (see Fig. \ref{fig:pipe}). This demonstrates a key advantage of \ac{REACT} that was not visible in the simulation study, where tether forces were not modeled. Although the \ac{REACT} adds computation compared to the baseline, the maximum replanning time of 0.4822~s remained well within the onboard platform’s capabilities, allowing real-time operation. This overhead, while higher than the conventional \ac{CPP} approach, was low enough to enable timely adaptation to tether entanglement events. These results demonstrate that standard planners without tether constraints fail even during relatively simple underwater 3D inspection tasks. While the baseline initially followed the planned path, it became entangled and subsequently immobilized midway through the mission. On the other hand, \ac{REACT}'s ability to detect and avoid entanglement through replanning enabled a successful mission completion. These findings highlight the need for entanglement-aware planning in real underwater operations.

\begin{table}[t]
    \centering
    \caption{Coverage Performance in Real-world Trials. \ac{REACT} takes a maximum of 0.4822 s (per re-planning step) during the inspection to compute the fully replanned path.}
    \label{tab:realworld_coverage}
    \resizebox{0.99\columnwidth}{!}{%
    \begin{tabular}{|l|c|c|c|}
        \hline
        \textbf{Planner} & \textbf{Inspection} & \textbf{ Coverage (\%)} & \textbf{ Computation overhead (s)} \\
        \hline
        \ac{REACT} & Full & 95.6 & 0.4822 \\
        CPP & Failed (Entangled) & 80.9 & N/A \\
        \hline
    \end{tabular}%
    }
    \vspace{0.5em}
\end{table}

\section{Conclusion}
\label{sec:conclusion}

We present \ac{REACT}, a real-time, entanglement-aware path planning framework for tethered underwater vehicles that supports underwater asset inspection tasks. Through comparative simulation and real-world experiments, we demonstrate that conventional planners face tether entanglement issues, rendering incomplete missions, as they cannot account for tether constraints. In simulation, \ac{REACT} shows effective tether management while maintaining full coverage, achieving a shorter total mission time ($680$s vs. $855$s) despite longer inspection time due to its proactive entanglement avoidance that eliminates the need for extensive post-mission disentanglement. The real-world experiments confirm these findings, where the baseline \ac{CPP} planner achieves only $80.9\%$ coverage before becoming physically stuck due to tether entanglement, while \ac{REACT} completes the full mission with $95.6\%$ coverage through adaptive replanning. 







\section*{Acknowledgment} 
 This work was supported by Innovation Fund Denmark (Grant No. 2040-00032B) and EIVA a/s. The authors thank Nikolas, Tom, and Leif from DFKI for experimental support.


\bibliographystyle{IEEEtran}
\bibliography{references}

@article{petit2022tape,
  title={Tape: Tether-aware path planning for autonomous exploration of unknown 3d cavities using a tangle-compatible tethered aerial robot},
  author={Petit, Louis and Desbiens, Alexis Lussier},
  journal={IEEE Robotics and Automation Letters},
  volume={7},
  number={4},
  pages={10550--10557},
  year={2022},
  publisher={IEEE}
}

@article{cao2023neptune,
  title={Neptune: nonentangling trajectory planning for multiple tethered unmanned vehicles},
  author={Cao, Muqing and Cao, Kun and Yuan, Shenghai and Nguyen, Thien-Minh and Xie, Lihua},
  journal={IEEE Transactions on Robotics},
  volume={39},
  number={4},
  pages={2786--2804},
  year={2023},
  publisher={IEEE}
}

@inproceedings{rov_mccammon,
  title={Planning and executing optimal non-entangling paths for tethered underwater vehicles},
  author={McCammon, Seth and Hollinger, Geoffrey A},
  booktitle={2017 IEEE International Conference on Robotics and Automation (ICRA)},
  pages={3040--3046},
  year={2017},
  organization={IEEE}
}

@article{bhattacharya2012topological,
  title={Topological constraints in search-based robot path planning},
  author={Bhattacharya, Subhrajit and Likhachev, Maxim and Kumar, Vijay},
  journal={Autonomous Robots},
  volume={33},
  pages={273--290},
  year={2012},
  publisher={Springer}
}

@article{definitions,
  title={Entanglement definitions for tethered robots: Exploration and analysis},
  author={Battocletti, Gianpietro and Boskos, Dimitris and Toli{\ae}, Domagoj and Palunko, Ivana and De Schutter, Bart},
  journal={IEEE access},
  year={2024},
  publisher={IEEE}
}

@inproceedings{martinez2021optimization,
  title={Optimization-based trajectory planning for tethered aerial robots},
  author={Martinez-Rozas, Sim{\'o}n and Alejo, David and Caballero, Fernando and Merino, Luis},
  booktitle={2021 IEEE International Conference on Robotics and Automation (ICRA)},
  pages={362--368},
  year={2021},
  organization={IEEE}
}

@article{zhang2019planning,
  title={Planning coordinated motions for tethered planar mobile robots},
  author={Zhang, Xu and Pham, Quang-Cuong},
  journal={Robotics and Autonomous Systems},
  volume={118},
  pages={189--203},
  year={2019},
  publisher={Elsevier}
}

@article{hert1996ties,
  title={The ties that bind: Motion planning for multiple tethered robots},
  author={Hert, Susan and Lumelsky, Vladimir},
  journal={Robotics and autonomous systems},
  volume={17},
  number={3},
  pages={187--215},
  year={1996},
  publisher={Elsevier}
}

@article{patil2023coordinating,
  title={Coordinating Tethered Autonomous Underwater Vehicles towards Entanglement-Free Navigation},
  author={Patil, Abhishek and Park, Myoungkuk and Bae, Jungyun},
  journal={Robotics},
  volume={12},
  number={3},
  pages={85},
  year={2023},
  publisher={MDPI}
}

@article{hert1999motion,
  title={Motion planning in R/sup 3/for multiple tethered robots},
  author={Hert, Susan and Lumelsky, Vladimir},
  journal={IEEE transactions on robotics and automation},
  volume={15},
  number={4},
  pages={623--639},
  year={1999},
  publisher={IEEE}
}

@article{cao2023path,
  title={Path planning for multiple tethered robots using topological braids},
  author={Cao, Muqing and Cao, Kun and Yuan, Shenghai and Liu, Kangcheng and Wong, Yan Loi and Xie, Lihua},
  journal={arXiv preprint arXiv:2305.00271},
  year={2023}
}

@inproceedings{mechsy2017novel,
  title={A novel offline coverage path planning algorithm for a tethered robot},
  author={Mechsy, LSR and Dias, MUB and Pragithmukar, W and Kulasekera, AL},
  booktitle={2017 17th International Conference on Control, Automation and Systems (ICCAS)},
  pages={218--223},
  year={2017},
  organization={IEEE}
}

@inproceedings{kim,
  title={Path planning for a tethered mobile robot},
  author={Kim, Soonkyum and Bhattacharya, Subhrajit and Kumar, Vijay},
  booktitle={2014 IEEE International Conference on Robotics and Automation (ICRA)},
  pages={1132--1139},
  year={2014},
  organization={IEEE}
}

@inproceedings{withy,
  title={WiTHy A*: Winding-Constrained Motion Planning for Tethered Robot using Hybrid A},
  author={Chipade, Vishnu S and Kumar, Rahul and Yong, Sze Zheng},
  booktitle={2024 IEEE International Conference on Robotics and Automation (ICRA)},
  pages={8771--8777},
  year={2024},
  organization={IEEE}
}

@inproceedings{bircher2015structural,
  title={Structural inspection path planning via iterative viewpoint resampling with application to aerial robotics},
  author={Bircher, Andreas and Alexis, Kostas and Burri, Michael and Oettershagen, Philipp and Omari, Sammy and Mantel, Thomas and Siegwart, Roland},
  booktitle={2015 IEEE International Conference on Robotics and Automation (ICRA)},
  pages={6423--6430},
  year={2015},
  organization={IEEE}
}

@inproceedings{feng2024fc,
  title={Fc-planner: A skeleton-guided planning framework for fast aerial coverage of complex 3d scenes},
  author={Feng, Chen and Li, Haojia and Zhang, Mingjie and Chen, Xinyi and Zhou, Boyu and Shen, Shaojie},
  booktitle={2024 IEEE International Conference on Robotics and Automation (ICRA)},
  pages={8686--8692},
  year={2024},
  organization={IEEE}
}

@article{dang2020graph,
  title={Graph-based subterranean exploration path planning using aerial and legged robots},
  author={Dang, Tung and Tranzatto, Marco and Khattak, Shehryar and Mascarich, Frank and Alexis, Kostas and Hutter, Marco},
  journal={Journal of Field Robotics},
  volume={37},
  number={8},
  pages={1363--1388},
  year={2020},
  publisher={Wiley Online Library}
}

@inproceedings{amer2023unav,
  title={Unav-sim: A visually realistic underwater robotics simulator and synthetic data-generation framework},
  author={Amer, Abdelhakim and {\'A}lvarez-Tu{\~n}{\'o}n, Olaya and U{\u{g}}urlu, Halil {\.I}brahim and Sejersen, Jonas Le Fevre and Brodskiy, Yury and Kayacan, Erdal},
  booktitle={2023 21st International Conference on Advanced Robotics (ICAR)},
  pages={570--576},
  year={2023},
  organization={IEEE}
}

@inproceedings{amer2023visual,
  title={Visual Tracking Nonlinear Model Predictive Control Method for Autonomous Wind Turbine Inspection},
  author={Amer, Abdelhakim and Mehndiratta, Mohit and le Fevre Sejersen, Jonas and Pham, Huy Xuan and Kayacan, Erdal},
  booktitle={2023 21st International Conference on Advanced Robotics (ICAR)},
  pages={431--438},
  year={2023},
  organization={IEEE}
}

@inproceedings{nvblox,
  title={nvblox: Gpu-accelerated incremental signed distance field mapping},
  author={Millane, Alexander and Oleynikova, Helen and Wirbel, Emilie and Steiner, Remo and Ramasamy, Vikram and Tingdahl, David and Siegwart, Roland},
  booktitle={2024 IEEE International Conference on Robotics and Automation (ICRA)},
  pages={2698--2705},
  year={2024},
  organization={IEEE}
}

@article{amergp,
  title={Empowering Autonomous Underwater Vehicles Using Learning-Based Model Predictive Control With Dynamic Forgetting Gaussian Processes},
  author={Amer, Abdelhakim and Mehndiratta, Mohit and Brodskiy, Yury and Kayacan, Erdal},
  journal={IEEE Transactions on Control Systems Technology},
  year={2025},
  publisher={IEEE}
}

@inproceedings{roperrt,
  title={RRT-Rope: A deterministic shortening approach for fast near-optimal path planning in large-scale uncluttered 3D environments},
  author={Petit, Louis and Desbiens, Alexis Lussier},
  booktitle={2021 IEEE International Conference on Systems, Man, and Cybernetics (SMC)},
  pages={1111--1118},
  year={2021},
  organization={IEEE}
}

@article{ompl,
  title={The open motion planning library},
  author={Sucan, Ioan A and Moll, Mark and Kavraki, Lydia E},
  journal={IEEE Robotics \& Automation Magazine},
  volume={19},
  number={4},
  pages={72--82},
  year={2012},
  publisher={IEEE}
}

@misc{amer2025modelling,
  title={Modelling of Underwater Vehicles using Physics-Informed Neural Networks with Control}, 
  author={Abdelhakim Amer and David Felsager and Yury Brodskiy and Andriy Sarabakha},
  year={2025},
  eprint={2504.20019},
  archivePrefix={arXiv},
  primaryClass={cs.RO},
  url={https://doi.org/10.48550/arXiv.2504.20019}
}

@article{peng2025spanning,
  title={Spanning-tree based coverage for a tethered robot},
  author={Peng, Xiao and Schwarzentruber, Fran{\c{c}}ois and Simonin, Olivier and Solnon, Christine},
  journal={IEEE Robotics and Automation Letters},
  year={2025},
  publisher={IEEE}
}

@article{teshnizi2021motion,
  author    = {Reza H. Teshnizi and Dylan A. Shell},
  title     = {Motion Planning for a Pair of Tethered Robots},
  journal   = {Autonomous Robots},
  volume    = {45},
  number    = {4},
  pages     = {503--521},
  year      = {2021},
  doi       = {10.1007/s10514-021-09972-x},
  url       = {https://doi.org/10.1007/s10514-021-09972-x}
}

@inproceedings{teshnizi2014computing,
  title={Computing cell-based decompositions dynamically for planning motions of tethered robots},
  author={Teshnizi, Reza H and Shell, Dylan A},
  booktitle={2014 IEEE International Conference on Robotics and Automation (ICRA)},
  pages={6130--6135},
  year={2014},
  organization={IEEE},
  doi={10.1109/ICRA.2014.6907762}
}

@phdthesis{amer2025autonomous,
  title={Autonomous Marine Robots Optimal Control and Path Planning with Data-Driven Models},
  author={Amer, Abdel Hakim Khaled Saad Amin},
  year={2025},
  school={AARHUS UNIVERSITY}
}
%

\end{document}